# Mining and Exploiting Domain-Specific Corpora in the PANACEA Platform

**Núria Bel[†], Vassilis Papavasiliou[*], Prokopis Prokopidis[*], Antonio Toral[‡], Victoria Arranz[§]**

[†]Universitat Pompeu Fabra, Barcelona, Spain
[*]Institute for Language and Speech Processing/Athena RIC, Athens, Greece
[‡]Dublin City University, Dublin, Ireland
[§]ELDA, Paris, France

E-mails: {nuria.bel}@upf.edu, {vpapa,prokopis}@ilsp.gr, {atoral}@computing.dcu.ie, {arranz}@elda.org

**Abstract**

The objective of the PANACEA ICT-2007.2.2 EU project is to build a platform that automates the stages involved in the acquisition, production, updating and maintenance of the large language resources required by, among others, MT systems. The development of a Corpus Acquisition Component (CAC) for extracting monolingual and bilingual data from the web is one of the most innovative building blocks of PANACEA. The CAC, which is the first stage in the PANACEA pipeline for building Language Resources, adopts an efficient and distributed methodology to crawl for web documents with rich textual content in specific languages and predefined domains. The CAC includes modules that can acquire parallel data from sites with in-domain content available in more than one language. In order to extrinsically evaluate the CAC methodology, we have conducted several experiments that used crawled parallel corpora for the identification and extraction of parallel sentences using sentence alignment. The corpora were then successfully used for domain adaptation of Machine Translation Systems.

**Keywords:** web crawling, boilerplate removal, corpus acquisition, IPR for language resources.

## 1. Introduction

There is a growing literature on using the Web for constructing large collections of monolingual and parallel collections. Such resources can be used by linguists studying language use and change (Kilgarriff and Grefenstette, 2003), and at the same time be exploited in applied research fields like machine translation, cross-lingual information retrieval, multilingual information extraction, etc. Moreover, these large collections of raw data can be automatically annotated and used to produce, by means of induction tools, a second order or synthesized derivatives: rich lexica (with morphological, syntactic and lexico-semantic information) and massive bilingual dictionaries (word and multiword based) and transfer grammars. The PANACEA LR factory is an interoperability platform of components creating complex workflows that reproduce the step-by-step process of creating such LRs. Facilities for searching, accessing web services as well as detailed documentation for each web service can be found at the PANACEA registry (http://registry.elda.org). Besides, the PANACEA myExperiment (http://myexperiment.elda.org) offers documentation, search and access facilities and a social platform for PANACEA workflows.

This paper focuses on web services and already deployed workflows for acquiring monolingual and bilingual domain-specific data. We also report on how they can be used for domain adaptation of Statistical Machine Translation (SMT) systems. The corpus acquisition procedure is described in Section 2. Further processing of the acquired data (i.e. sentence extraction, sentence alignment, etc) and their exploitation for domain adaptation are presented in Section 3. In Section 4, issues concerning the Intellectual Property Rights of the produced resources are discussed. Conclusions and future work are reported in Section 5.

## 2. Corpus Acquisition

In order to construct large-scale domain-specific collections, we developed a Corpus Acquisition Component (CAC) which consists of a focused monolingual (FMC) and a focused bilingual crawler (FBC). Both crawlers have been deployed as web services in the PANACEA platform and are available at http://nlp.ilsp.gr/soaplab2-axis/.

### 2.1 Acquiring monolingual data

The FMC adopts a distributed computing architecture based on Bixo[1] (an open source web mining toolkit that runs on top of Hadoop[2]) and integrates modules for parsing web pages, text normalization, language identification, document clean-up and text classification. A required input resource from the user is a description of the targeted topic in a specific language. For topic description, we adopted a strategy proposed by Ardö and Golub, (2007) i.e. using triplets (`<term, relevance weight, topic-class>`) as the basic entities of the topic definition[3]. Topic definitions can be constructed manually or by repurposing online resources like the Eurovoc multilingual thesaurus that we used during development. Another required input is a list of seed URLs pointing to a few relevant web pages that are used to initialize the crawler.

---

[1] http://openbixo.org/
[2] http://hadoop.apache.org/

Each fetched web page is parsed in order to extract its metadata and content. Then, the content is converted into a unified text encoding (UTF-8) and analyzed by the embedded language identifier. If the document is not in the targeted language, it is discarded. In addition, the language identifier is applied at paragraph level and paragraphs in a language other than the main document language are marked as such.

Next, each crawled, normalized and in-target language web page is compared with the topic definition. Based on the amount of terms' occurrences, their locations (i.e. title, keywords, body), and their weights, a relevance score is estimated. If this value exceeds a predefined threshold, the web page is classified as relevant and stored.

Relevant or not, each web page is parsed and its links are extracted and prioritized according to a) the relevance-to-the-topic score of their surrounding text and b) the relevance-to-the-topic score of the web page they were extracted from. Following the most promising links, FMC visits new pages and continues until a termination criterion is satisfied (i.e. time limit).

In order to provide corpora useful for linguistic purposes, FMC employs the Boilerpipe tool (Kohlschütter et al., 2010) to detect and mark parts of the HTML source that are usually redundant (i.e. advertisements, disclaimers, etc).

The final output of the FMC is a set of XML documents following the Corpus Encoding Standard[4]. An XML file relevant to the Environment domain in French can be found at http://nlp.ilsp.gr/nlp/examples/2547.xml.

### 4.1 Acquiring bilingual data

The FBC integrates the FMC and a module for detecting pairs of parallel documents. The required input from the user consists of a list of terms that describe a topic in two languages and a URL pointing to a multilingual web site. The FBC starts from this URL and in a spider-like mode extracts links to pages inside the same web site. Extracted links are prioritized according to the probability that they point to a translation of the web page they originated from, and the two criteria mentioned in 2.1. Following the most promising links, FBC keeps visiting new pages from the web site until no more links can be extracted.

After this stage, the pair detection module, inspired by Bitextor (Esplà-Gomis and Forcada, 2010), examines the structure of the downloaded pages to identify pairs of parallel documents. The module performs better on document pools from well-organized web sites, i.e. multilingual sites with pages containing links to translations comparable in structure and length. The final output of the FBC is a list of XML files, each pointing to a pair of files in the targeted languages[5].

## 3. Alignment

At this stage, we have pairs of documents produced by the FBC (see Section 2.2). In order to take advantage of this data, it should be aligned at a finer level, i.e. sentence alignment and word alignment. PANACEA has developed web services for a set of state-of-the-art sentence and word aligners. Namely, for sentence alignment, we provide web services for Hunalign, BSA and GMA. Regarding word alignment, GIZA++, Berkeley Aligner and Anymalign have been integrated. All these web services are available at http://www.cngl.ie/panacea-soaplab2-axis/. For a more detailed description of alignment web services, their implementation and deployment, please refer to (Toral et al., 2011).

The sentence-aligned data can then be used for a variety of tasks. For example, we have used this kind of data to adapt a Statistical Machine Translation system to given specific domains (environment and labour legislation) and language pairs (English--French and English--Greek) (Pecina et al., 2011). By using the domain-specific crawled and sentence-aligned data, we are able to improve the performance of Machine Translation by up to 48%.

Another use is the production of domain-specific Translation Memories (Poch et al., 2012). In this case, the data received from the FBC is first sentence-aligned and then converted into TMX, the most common format used to encode Translation Memories. This is deemed to be very useful for translators when they start translating documents for a new domain. As at that early stage they still do not have any content in their TM, having the automatically acquired TM can be helpful in order to get familiar with the characteristic bilingual terminology and other aspects of the domain.

## 4. IPR case study

It is the aim of PANACEA to explore all the issues related to the usability of the produced resources. Thus, work on the exploitation plan of PANACEA has led to an interesting study of the type of assets produced. This project offers a combination of data, software and web services that need to be considered at different stages. Here we focus on the specific work done on the monolingual and bilingual data described in the previous sections with the aim to establishing an appropriate and clear legal framework for its exploitation in all possible scenarios. Given the trend nowadays to crawl and use data from internet, we considered this case study as crucial for this and other similar data-production approaches in particular seeing current initiatives choose options like leaving IPR issues in-handled, in the hands of the future users themselves or praising for the good nature of the owners who may take them to court. Once the internet data to be used has been listed, the procedure followed to clear out their IPR issues follows these steps: - Locating all sources and contact points. - Studying terms and conditions (use and possible distribution, if any). - Approaching providers (mostly, on a case per case basis). The complexity behind this procedure ranges considerably due to factors such as: source type; access to some institutions and blogs; need to reassure sources of no ownership right infringement; need to explain data use to data users (what is HLT?); data size; allowed negotiation time (generally long but where the needs of the future data users impose some clear restrictions). In the case of PANACEA, a large number of URLs were to be handled, 14,479, which contained 190,540 pages as a whole. However, given the cost and time restrictions imposed by both the task and the project budget, only the most frequent URLs were selected to undergo negotiation. Thresholds were set up as follows:

- For monolingual data: after an initial collection of relatively small corpora (which was performed early in the project and resulted in storing 5,623 pages from 1,175 web sites), web sites, for which under 7 pages were collected, were not examined for IPR issues; after a second experiment, which resulted in a much larger collection (184,917 pages from 13,304 sites), web sites with under 100 pages were not considered
- For bilingual and aligned data: all sources were targeted. IPR clearance was given top priority given their processing effort (aligned). 27 URLs were contained in both batches of the bilingual data, with their respective 1,948 pages. An interesting conclusion of this work was the analysis of negotiation duration and status reached at this stage of the project (with Year 2 already completed). Leaving aside refused negotiations (e.g., already 10 in the case of the monolingual data), which is a fact that should not be neglected for similar approaches, monolingual negotiations have taken between 1 day (for very fast replies) to 339 days, which shows and average duration of 66 days. Bilingual data have taken between 8 to 344 days, with an average of 176 days. This seems to be the hard reality to be faced when aiming to handle IPR for such type of data.

---

[3] http://nlp.ilsp.gr/panacea/testinput/monolingual/ENV_topics/ENV_EN_topic.txt is an example of a list of English terms for the environment.

[4] http://www.xces.org/

[5] http://nlp.ilsp.gr/panacea/xces-xslt/202_225.xml links a pair of documents in English and Greek

## 5. Conclusions

PANACEA project is working for the automatic production of language resources that are the critical components for the multilingual, domain tuned applications embodying different language technologies. We have presented the services already available to produce usable domain-specific aligned corpora based on the parallel data found in the internet. These web services can be chained in workflows that implement the project goals: the automatic production of language resources. By the end of the year, PANACEA will also offer web services and workflows for the automatic production of other resources such as bilingual dictionaries, monolingual rich lexica containing verb subcategorization frame information, selectional preferences, multiword extraction, and lexical semantic class of nouns.